\documentclass[letterpaper, 10 pt, conference]{ieeeconf}  

\IEEEoverridecommandlockouts                              

\usepackage{cite}
\usepackage{algorithmic}
\usepackage{graphicx}
\usepackage{xcolor}
\usepackage{url}
\usepackage{algorithm}
\usepackage{subfigure}
\usepackage{svg}
\usepackage{ulem}

\title{\LARGE \bf
coExplore: Combining multiple rankings for multi-robot exploration
}

\author{Ingo Scheler$^{1}$, Robin Dietrich$^{1}$
\thanks{$^{1}$Department of Informatics,
        Technical University of Munich; Garching, 85748 Germany;
        {e-mail: ingo.scheler@tum.de,
                    robin.dietrich@tum.de
        }
        $^{2}$ https://www.github.com/Icheler/coExplore
        }%
}

\begin{document}

\maketitle
\thispagestyle{empty}
\pagestyle{empty}

\begin{abstract}

Multi-robot exploration is a field which tackles the challenge of exploring a previously unknown environment with a number of robots. This is especially relevant for search and rescue operations where time is essential. Current state of the art approaches are able to explore a given environment with a large number of robots by assigning them to frontiers. However, this assignment generally favors large frontiers and hence omits potentially valuable medium-sized frontiers. 

In this paper we showcase a novel multi-robot exploration algorithm, which improves and adapts the existing approaches. Through the addition of information gain based ranking we improve the exploration time for closed urban environments while maintaining similar exploration performance compared to the state-of-the-art for open environments. Accompanying this paper, we further publish our research code in order to lower the barrier to entry for further multi-robot exploration research. We evaluate the performance in three simulated scenarios, two urban and one open scenario, where our algorithm outperforms the state of the art by 5\% overall.

\end{abstract}


\section{Introduction}\label{chapter:introduction}

\begin{figure*}
\centerline{\subfigure[]{\includegraphics[width=0.3\linewidth]{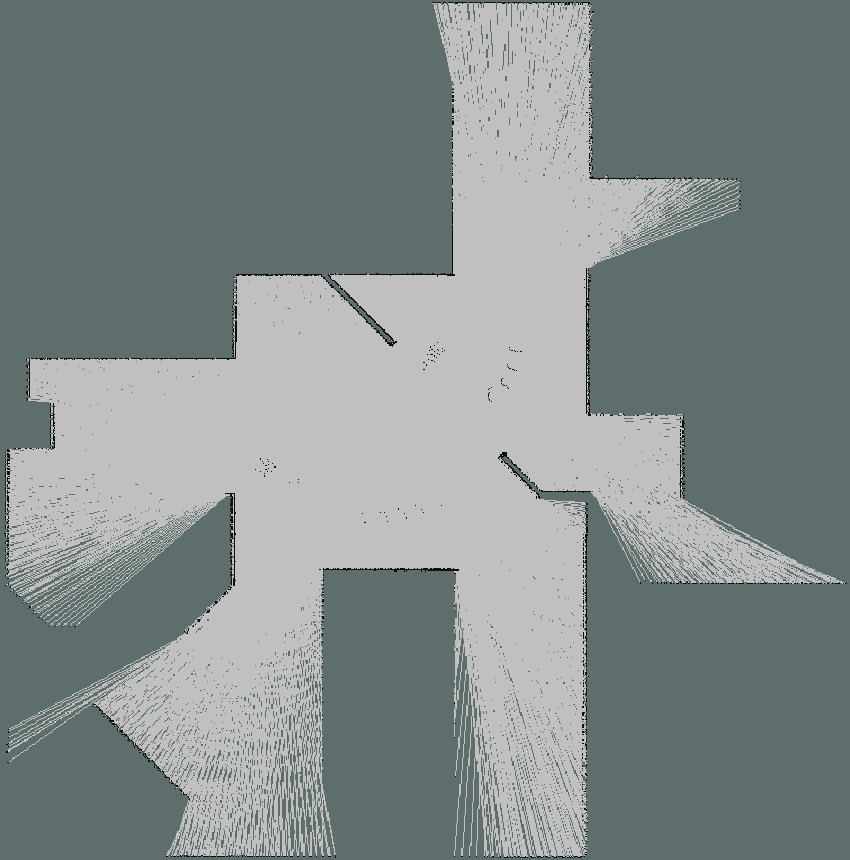}
\label{fig:plainmap}}
\hfil
\subfigure[]{\includegraphics[width=0.3\linewidth]{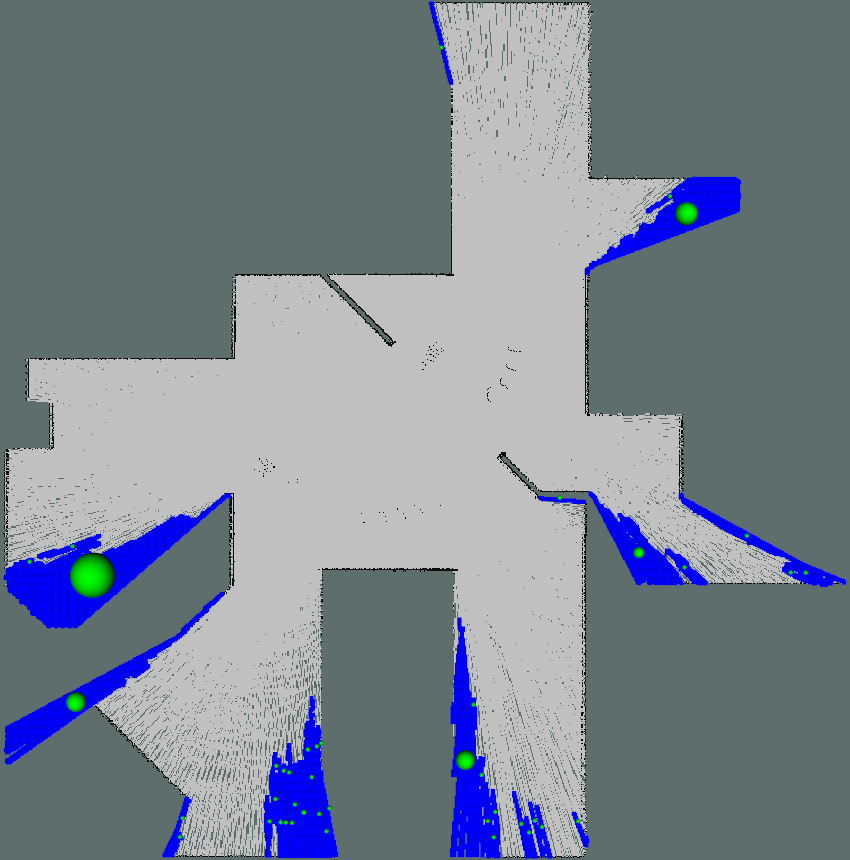}
\label{fig:frontiermap}}
}
\caption{A global grid-map obtained by joining the output of all individual SLAM algorithms (a) and the map with identified frontiers (blue) as well as their respective sizes (green) (b).
}
\label{fig:maps}
\end{figure*}

Search and Rescue operations operate under the assumption, that the longer it takes to find lost people, the less likely their survival is \cite{adams_search_2007}, \cite{noauthor_search_2020}. This emphasises the need for fast exploration in unknown environments. While singular robot systems can achieve the task of exploration, the addition of multiple robots can greatly decrease the time needed to explore a given environment, since the robots are able to distribute themselves across the area and explore more of it earlier on during the exploration \cite{burgard_coordinated_2005}. Distributing the robots in a near optimal way to ensure fast and efficient exploration is a non-trivial problem however.

The field of robotic exploration works with map data, as shown in Fig. \ref{fig:plainmap}, obtained by robots sensing their surroundings through sensors like lidars or cameras. Based on this map data, points in the environment associated with edges between unknown and explored space are identified, so called \textit{frontiers}. This is demonstrated in Fig. \ref{fig:frontiermap} where the detected frontiers, edges between explored and unknown space, are highlighted in blue and their size is visible in green. Algorithms in multi-robot exploration distribute robots to these frontiers to gain more information about the environment and explore them according to pre-defined rules. These are defined depending on which metrics are evaluated, such as overall exploration time needed to explore a given environment. While the first frontier exploration technique only assigned robots to the nearest frontier \cite{yamauchi_frontier-based_1997}, subsequent techniques further optimized the approach by assigning robots to frontiers based on heuristics \cite{burgard_coordinated_2005}, and assigning ranks based on the distance from each frontier to each robot \cite{bautin_minpos_2012}. Metrics like distance from robot to frontiers and information gain, i.e. the factor of how much new information we expect to gain by exploring a frontier, have been used in a variety of methods, like in Colares et al. \cite{colares2016next}. MinPos \cite{bautin_minpos_2012} uses the frontier distance to decide if robots have the same rank or Burgard et al.\cite{burgard_coordinated_2005} which uses information gain to find good frontiers and minimizing overlap between the explored map sections. We have identified two problems with the mentioned algorithms. First Yamauchi et al \cite{yamauchi_frontier-based_1997} and minPos \cite{bautin_minpos_2012} do not make use of information gain. Burgard et al and nextFrontier \cite{colares2016next} rely too heavily on this information which hinders exploration time by only heading for the currently best points, which in terms of information gain and the long-term value of these frontiers are local minima.

For this we propose a new algorithm which utilizes assignment through the use of the Hungarian algorithm \cite{kuhn_hungarian_1955}, this minimizes the cost matrix composed of robot distances to frontiers, rank of frontiers like in \cite{bautin_minpos_2012} and rank of possible information gain from frontiers. Furthermore accompanying this paper we provide an open-source evaluation tool$^{2}$ for fast prototyping of metrics, distance measurement and assignment methods to lower the barrier to entry in multi-robot exploration. This tool is used in our own evaluation, demonstrating the validity of our proposed approach against the current state of the art algorithms. We demonstrate how our approach, an extension to minPos \cite{bautin_minpos_2012}, outperforms previous methods in all tested scenarios, 2 structured and one open environment, by 5\% and performs especially well for structured environments where we outperform by 6\%.

Over the course of the paper we show how current approaches in the field of multi-robot exploration work in section \ref{chapter:related}. Then we given an explanation of our method in chapter \ref{chapter:approach} and review how our algorithm compares to the previously explained competition and state of the art in section \ref{chapter:evaluation}. The paper is concluded with an overview of the presented information in chapter \ref{chapter: conclusion} followed by insights into how this work can be improved upon in the future in the final section \ref{chapter: future work}.

\section{Related work}\label{chapter:related}
In 1997 with the novel introduction of frontier-based exploration by Yamauchi et al \cite{yamauchi_frontier-based_1997}, multi-robot exploration got its first important metric. Emphasizing a simple strategy to explore an environment while moving as little as possible. Robots explore their nearest frontier until no further frontiers can be explored. The approach was originally proposed for the single robot case, then it was further adapted to the multi-robot case by using the nearest frontier strategy for all robots in the scenario while sharing map data \cite{yamauchi_decentralized_1999}. This approach is also the first decentralized multi-robot exploration approach where every robot runs their own allocation algorithm which prevents failure of the system if robots fail due to software errors or damage. This has set the groundwork for all further approaches which optimize for different metrics to further and more quickly explore given environments. The approach can be referred to as \textit{nearest} in further mentions.

The next notable approach was proposed by Burgard et al.\cite{burgard_coordinated_2005} which makes use of the expected information gain of a given frontier to match robots and frontiers with each other while maximizing the overall possible information gained by sending a given robot to a specific frontier. This metric can be computed in different ways. Burgard et al.\cite{burgard_coordinated_2005} chose to determine a frontiers utility by projecting the robot onto that point and determining how many cells that robot would spot given the sensor used for vision. Then the robots would be incrementally assigned to available frontiers while spreading the robots, so a maximum of possible cells would be spotted.

Simpler versions of the visibility metric have been used in current approaches like nextFrontier \cite{colares2016next}, which computes for each frontier cell how many unknown cells are adjacent and determines a utility for each frontier. Additionally they use a novel way to compute the distance utility of each frontier by favoring frontiers which are neither closest nor furthest away. Lastly they use a reverse distance metric to find frontiers in the environment which are furthest away from all other available robots. These metrics are combined during assignment to find the best possible frontier to explore for each robot.

\label{approach:minPos}
MinPos \cite{bautin_minpos_2012} finds the distance between frontiers and robots by wavefront propagation \cite{Barraquand1991}. A rank is assigned based on the relative position of robots to frontiers, i.e. the closest robot to a frontier gets the lowest rank increasing to the maximal rank assigned if the robot is the furthest away. If multiple frontiers have the lowest rank for a given robot, the closest frontier is chosen. This approach emphasizes the spread of robots in the environment. Similarly to the nearest frontier exploration \cite{yamauchi_decentralized_1999} it minimizes the travel time of a robot, while further improving exploration time by finding a better spatial distribution of robots in the environment. As with the nearest approach, the algorithm does not utilize the possible information gain of frontiers. Variations of this approach have been proposed, such as the one by Dileep et al. \cite{dileep_muddu_frontier_2015}, where frontiers are marked for exploration if they are within a certain distance threshold to the robot. Then before new frontiers are explored all previously marked frontiers are explored first. The performance of this approach was, however, slightly worse compared to the base algorithm. 

All of the introduced approaches exist as decentralized and centralized versions. Normally multi-robot exploration hedges against single robot failure, where the failure of a single part of the system does not impact the overall system more than the lost utility of the lost robot. This problem is reintroduced by using a single system to compute the robot distribution. If the gathering system fails, the assignment for all robots is hindered until a new monolith is decided upon. Decentralized approaches on the other hand only compute the combined map of all partial maps of the robots by communication and use different strategies like auction based distribution to decide the robot distribution to frontiers in a decentralized manner \cite{hussein_multi-robot_2014}. For all previous papers \cite{yamauchi_decentralized_1999}, \cite{burgard_coordinated_2005}, \cite{colares2016next}, \cite{bautin_minpos_2012} there exist versions which utilize decentralized approaches.

In our approach we combine metrics from previous works in order to lower the exploration time. For this reason we use the distance from robots to frontiers\cite{bautin_minpos_2012}, rank of robots based on distance to frontiers\cite{bautin_minpos_2012} and information gain\cite{colares2016next}. The combined metric is used during the assignment for which we chose the hungarian algorithm based on previous work done by Faigl et al.\cite{faigl_comparison_2015}. The authors evaluated multiple methods like multiple traveling salesman assignment, iterative assignment and hungarian assignment and demonstrate that the hungarian algorithm is competitive to other methods, such as iterative assignment. This method loops over all robots and chooses the frontier which is best for the current robot, removes that frontier and iteratively assigns the frontiers. This can lead to sub-optimal assignments over the complete set of frontiers and robots.

Through this combination of previous research we further reduce the exploration time needed during search and rescue operations. How this is achieved specifically, is explained in the following section.

\section{Co-Explore}\label{chapter:approach}

Our approach uses 2D Occupancy grid maps \cite{elfes_using_1989} to provide a representation of the environment which has to be explored, similar to approaches like minPos \cite{bautin_minpos_2012}. Occupancy grids provide a probabilistic representation of the environment by assigning values to each cell or pixel. This value depends on whether the cell has been explored, whether it is deemed empty or the likelihood that sensor readings estimate an obstacle to be in that cell. The grid map serves as the base for frontier extraction.

Each robot $r \in R$ generates its own grid map by running a separate SLAM algorithm \cite{macenski2021slam} to find its position and simultaneously map the environment. Afterwards, this data is merged together to obtain a complete map of the environment based on all individual maps of the robots ($R$). We use the \textit{multirobot\_map\_merge} package for this purpose, which was developed by Hoerner et al. \cite{Horner2016}. This process allows the localization of all robots $R$ within the global map and is a prerequisite for our method. Based on this information we aim to find frontiers in the environment, as can be seen in Fig. \ref{fig:maps}. Here, Fig. \ref{fig:plainmap} shows the data which we are provided with before assigning frontiers $F$ and sizes $S$, while Fig. \ref{fig:frontiermap} displays the map with completed frontier identification and sizing, for this we adapted the process of Hoerner et al. \cite{Horner2016}.

Upon completion of the frontier identification we use the data, i.e. robots $R$ with positions and information on frontiers ($F, S$), to start our pipeline leading to the final output - the assignment of each robot $r$ to one frontier position $f \in F$. The input is visible in Alg. \ref{alg: main}, which is then used in subsequent steps to compute all the information used by the Hungarian algorithm \cite{kuhn_hungarian_1955} for assignment. Firstly, we compute the distance $d \in D$ from each robot $r$ to each frontier $f$ through wavefront propagation \cite{Barraquand1991} see line 4 in Alg. \ref{alg: main}. For each frontier $f$, we then rank the distance to each robot $r$ ($Ra$), as in minPos \cite{bautin_minpos_2012} see line 7 in Alg. \ref{alg: main}. Finally, we rank frontiers $f$ in descending order ($Rs$), based on the number of cells associated with each of them ($S$), see line 10 in Alg. \ref{alg: main}. This process is described in more detail in the following subsection.

\subsection{Frontier identification and sizing}
In order to define new frontiers $f$ and determine their size $s \in S$, we identify the space, where unassigned parts in the map are adjacent to free space, adapted from Horner et al. \cite{Horner2016}. This is achieved by looking in the neighborhood of cells at the border to find all cells which satisfy the constraint of being at a border. These cells are then grouped together based on their immediate neighborhood and propagating outward from the first found cell until no new cells can be found in the vicinity. All cell locations are saved and based on the grouped cells, new frontiers are defined. Their center is determined by finding the mean position $f$ of the cluster and their overall size $s$ corresponds to the number of cells. With this information we can proceed to compute different metrics for the assignment. The result of this process is highlighted in Fig. \ref{fig:frontiermap}.

\begin{figure*}
\centerline{\subfigure[Maze-like environment]{\includegraphics[width=0.25\linewidth]{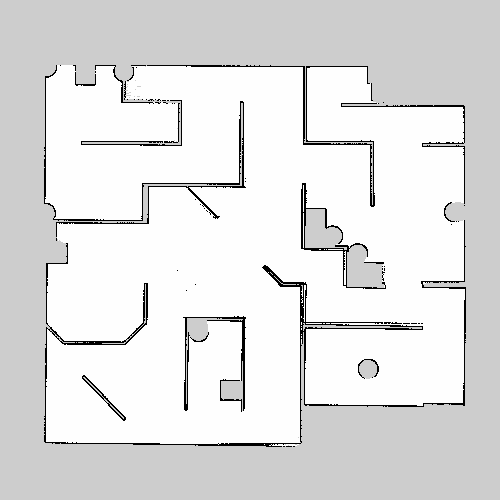}
\label{fig:mazeenv}}
\hfil
\subfigure[Office-like environment]{\includegraphics[width=0.25\linewidth]{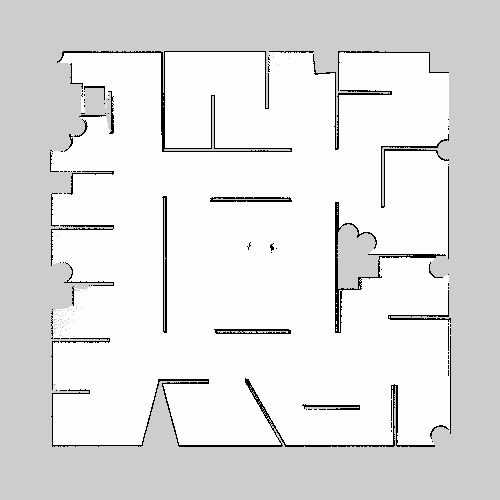}
\label{fig:officeenv}}
\hfil
\subfigure[Open environment]{\includegraphics[width=0.25\linewidth]{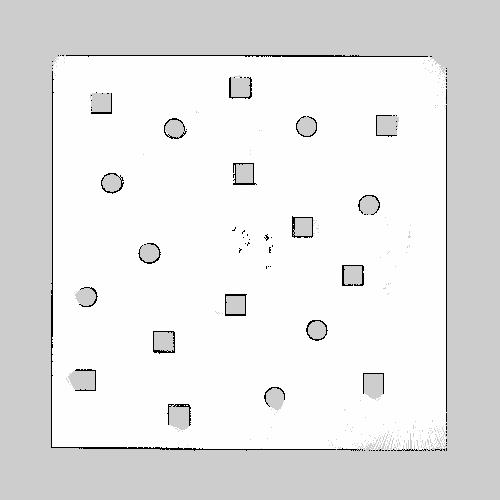}
\label{fig:openenv}}
}
\caption{Global grid maps of the respective environments after completing the exploration with 2 robots.}
\label{fig:environments}
\end{figure*}

\subsection{Metric computation}
\label{sub:metric}
\begin{algorithm}
\caption{coExplore and \textcolor{blue}{co122 in blue} run at 0.5 Hz}
\label{alg: main}
\begin{algorithmic}[1]
\REQUIRE Robots $R$, Frontiers with position $F$ and size $S$
    \STATE // \textcolor{orange}{Matrices} and \textcolor{teal}{Vectors} are highlighted
\REPEAT
    \STATE \textcolor{gray}{{\textit{// Calculate distances from robots to frontiers}}}
    \STATE $\textcolor{orange}{D} = distance(\textcolor{teal}{R},\textcolor{teal}{F})$
    \STATE \textcolor{gray}{\textit{// Calculate robot rank based on proximity to frontiers}}
    \FORALL{$f_j \in \textcolor{teal}{F}$}
        \STATE $\textcolor{teal}{Ra_j} = $ rankRobots$( \textcolor{teal}{D_j})$ \textcolor{gray}{// see section \ref{sub:metric}}
    \ENDFOR \\
    \STATE \textcolor{gray}{\textit{// Calculate frontier sizes and rank them}}
    \STATE $\textcolor{teal}{Fs} = $ rankSizes$(\textcolor{teal}{S})$  \textcolor{gray}{// see section \ref{sub:metric}}
    \FORALL{$r_i \in \textcolor{teal}{R}$}
        \STATE $\textcolor{orange}{Rs_i} = \textcolor{teal}{Fs}$
    \ENDFOR
    \STATE $\textcolor{orange}{Ra} = \textcolor{orange}{Ra} / max(\textcolor{orange}{Ra})$ \textcolor{gray}{// normalize}
    \STATE $\textcolor{orange}{D} = \textcolor{orange}{D} / max(\textcolor{orange}{D})$ \textcolor{gray}{// normalize}
    \STATE $\textcolor{orange}{Rs} = \textcolor{orange}{Rs} / max(\textcolor{orange}{Rs})$ \textcolor{gray}{// normalize}
    \STATE \textcolor{gray}{\textit{// Calculate assignments}}
    \STATE $\textcolor{orange}{X} = \textcolor{blue}{2 \cdot} \textcolor{orange}{Rs} + \textcolor{orange}{Ra} + \textcolor{blue}{2 \cdot}\textcolor{orange}{D}$
    \STATE $\textcolor{orange}{X} =$ appendHighValues(\textcolor{orange}{X}) \textcolor{gray}{// see section \ref{sub:assign}}
    \STATE \textcolor{teal}{optimal\_frontier\_indices} = hungarian(\textcolor{orange}{X}) \cite{kuhn_hungarian_1955}
    \STATE \textcolor{gray}{\textit{// Send robots to optimal frontiers}}
    \FORALL{$r_i \in \textcolor{teal}{R}$}
        \IF{\textcolor{teal}{optimal\_frontier\_indices$_i$}$ < |\textcolor{teal}{F}|$}
            \STATE sendRobotTo(\textcolor{teal}{optimal\_frontier\_indices$_i$}, $r_i$)
        \ELSE
            \STATE sendRobotTo(random($\textcolor{teal}{F}$), $r_i$)
        \ENDIF
    \ENDFOR
\UNTIL{$|F| = 0$}
\end{algorithmic}
\end{algorithm}

With the map information, the position of each robot $r$, and the frontier $f$ and size $s$ information, we can compute different features or metrics to combine later on. Firstly, the distance $d$ from each robot $r$ to each frontier $f$ is computed. The simplest method uses the euclidean distance between robot positions and each frontier. Other options include the use of wavefront propagation \cite{Barraquand1991} to plan waves from the robot position until all frontiers are reached to find distance measurements to frontiers based on the map data \cite{bautin_minpos_2012}. This propagation is used for all algorithms in our evaluation including our own versions. Wave propagation allows the robot to get approximate distances in the environment while taking into account obstacles in the environment, which euclidean distance omits. This is visible in Alg. \ref{alg: main} in lines 3-4. 

Next, we rank two different features, distances $Ra$ and frontier sizes $Rs$. This is shown in Alg. \ref{alg: main} in lines 5-13. For every frontier $f$ we compute the distance to each robot $r$ and rank the robots according to their proximity to the frontier ($Ra$), where the closest robot gets assigned the lowest rank ascending to the highest rank for the furthest robot starting with integers from 0, similar to \cite{bautin_minpos_2012}. Then we rank all frontiers based on their size ($Rs$). We invert the measurement, so the frontier with the biggest size has the smallest rank in our calculations. While the distance ranks allow us to spread the robots in the environment the theory behind the size rankings is different. For the simple method of looking at the size of frontiers $s$ and the advanced method of nextFrontier \cite{colares2016next}, the biggest problem lies in frontiers which are assumed to be very large. This is a result from Lidar sensors shooting out rays which cover 100 percent of the area close to the robot but their coverage decreases the farther away the point is from the robot. This is visible on the left hand side of both images in Fig. \ref{fig:maps}. The rays pierce through the unknown space and frontier identification methods assume the found frontier $f$ to be very large in size $s$. There are two options to fix this issue, either we rank $F$ so the impact of these measurements is reduced or we use visual computation to find the 'true' frontier edges for coarse frontiers. For our approach we chose the simpler solution to limit the computational complexity.

With $D, Ra$ and $Rs$ computed, we give each one a similar impact by normalizing them to a range from 0 to 1, where 0 has the biggest impact given the metric and 1 the least. We compute this by dividing the computed values by the maximum value in the associated matrix. This creates a small bias towards traveling based on distance to frontier since we only divide by the maximum without removing the minimum, which includes this inherent bias, since the other two metrics are exactly between 0 and 1. These normalizations are visible in Alg. \ref{alg: main} in lines 14-16.

Alg. \ref{alg: main} contains two algorithms, \textit{coExplore} and \textit{coExplore-122}, to which we later refer to as \textit{co122}. While coExplore embodies our base idea, co122 places an increased attention on the nearest distanced frontiers and the rank of frontiers. The difference is highlighted in line 18 of Alg. \ref{alg: main}. This leads to a more consistent performance in structured environments. The parameters for co122 were chosen during an exploratory internal benchmark of the impact of different weightings on exploration time, where that weighting showed the best performance in structured environments while increasing the emphasis on frontier sizes $S$ improved the performance in open environments.

\subsection{Summation and Assignment}
\label{sub:assign}
To find the optimal assignment of $R$ to $F$ we sum up the previous metrics ($D, Ra, Rs$) to $X$: distances to frontiers $D$, ranks based on these distances $Ra$ and ranks for the information gain per frontier $Rs$ as in Alg. \ref{alg: main} and line 18. Assignment methods were extensively tested for multi-robot exploration in Faigl et al.\cite{faigl_benchmarking_2015}. Then we use the Hungarian Algorithm \cite{kuhn_hungarian_1955} to find the optimal assignment by minimizing the overall cost, visible in line 20 of Alg. \ref{alg: main}. To allow the handling of the scenario where there are more robots than frontiers, we append the computed matrix $X$ with additional mock-frontiers with extremely large values which only get chosen in the described scenario. This allows us to always assign the optimal robots for the given frontiers while randomly assigning the remaining robots to the existing frontiers, since we assume that new frontiers can lead to newly explorable environments. This addition is visible in line 19 of Alg. \ref{alg: main}. The algorithm, as shown in Alg. \ref{alg: main}, is run at 0.5Hz without waiting for robots to reach frontiers. This frequency allows the robots to travel to their chosen destinations for sufficient amount of time while maintaining that the best frontiers are chosen every 2 seconds. We avoid oscillations between frontiers, since the travelling towards a frontier will favor it further. This way, newly gained information from robots traveling through the environment is continuously incorporated and robots are always assigned to the frontier which is optimal given the current information.

\section{Evaluation}
In this section we present the evaluation results of our algorithm in three different environments.
\label{chapter: experiments}\label{chapter:evaluation}
\subsection{Experimentation Setup}
All subsequently presented experiments were conducted in ROS1 noetic \cite{quigley2009ros} using Gazebo \cite{Gazebo2004} as  simulation environment. We simulated multiple Festo Robotino robots (Version 1) in Gazebo with a front mounted 2d Hokuyo Lidar with 10.6 meters range and 240 degrees FoV. These robots are interchangeable, the robot type does not impact the quality of our algorithmic solution, only the Lidar sensor and its properties does. The results were validated by using our self-developed ROS1 package in conjunction with SLAM capabilities provided by the \textit{slam\_toolbox} package \cite{macenski2021slam} as well as path planning and control functions by the \textit{move\_base} package \cite{lu_move_base_nodate}. The exploration times for each scenario, robot, map and algorithm, can vary depending on mapping and control quality, to minimize their impact we completed 15 runs per configuration. Due to computational limitations with the chosen simulation environments and packages leading to a degradation of performance for higher robot numbers, we are limited to testing with up to 5 robots at the same time.

We capture data on how far the robots have traveled and how much of the area has been covered. This information is collected and stored every 5 seconds. Based on this data we can evaluate different metrics, like map coverage over time, distance traveled over time and the difference between traveled distances of the robots. 

\subsection{Environments}
\begin{figure}
    \centering
    \includegraphics[width=0.5\textwidth]{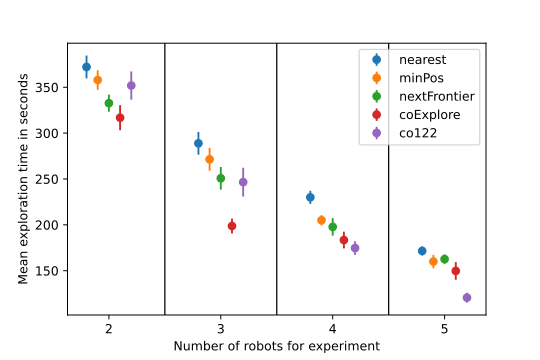}
    \caption{The mean exploration time and standard deviation after 15 runs in the maze environment (Fig. \ref{fig:mazeenv}) for $2$-$5$ robots.}
    \label{fig:maze_results}
\end{figure}
\begin{figure}
    \centering
    \includegraphics[width=0.5\textwidth]{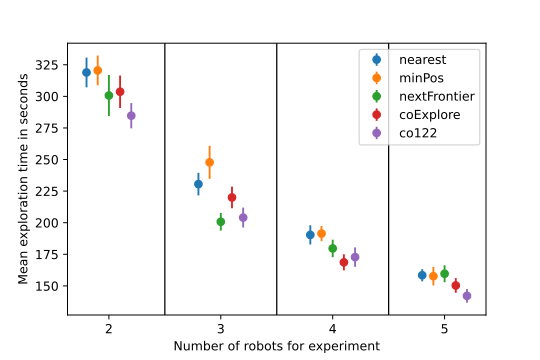}
    \caption{The mean exploration time and standard deviation after 15 runs in the office environment (Fig. \ref{fig:officeenv} for $2$-$5$ robots.}
    \label{fig:office_results}
\end{figure}
\begin{figure}
    \centering
    \includegraphics[width=0.5\textwidth]{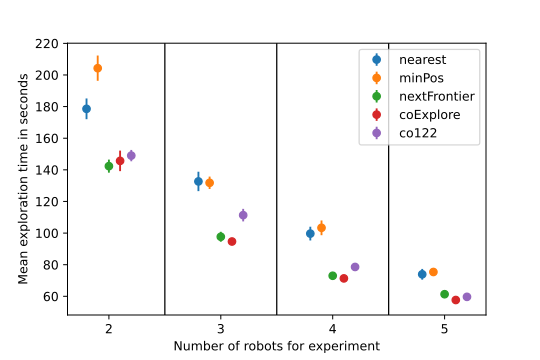}
    \caption{The mean exploration time and standard deviation after 15 runs in the open environment \ref{fig:openenv} for $2$-$5$ robots.}
    \label{fig:open_results}
\end{figure}

We validate our proposed approach through simulation with ROS in 3 distinct environments, visible in Fig. \ref{fig:mazeenv}, \ref{fig:officeenv} and \ref{fig:openenv}. Two of them are more urban or structured environments whereas the last one is an open environment modeled after the \textit{potholes} environment of a multi-robot exploration benchmark \cite{faigl_benchmarking_2015}, including an open space with multiple small objects to allow for robust SLAM. The environment and maps used for the benchmark are not publicly available, therefore we approximated the open environment as close as possible with our available resources. All environments have unique features which enable the testing of the algorithms with different focuses. The maze environment (Fig. \ref{fig:mazeenv}) has a small subset of explorable areas, which need large travel times to fully discover. This heavily penalizes the constant reassignment of new frontiers, for example when looking primarily at the possible information gain. The next environment (Fig. \ref{fig:officeenv}) has a large amount of small areas which need to be explored and are accessible through a variety of paths. This is modeled after a simplified office environment to highlight how algorithms evaluate a great set of frontiers in an open environment, where the paths are not restricted as they are in the map shown in Fig. \ref{fig:mazeenv}. The open environment (Fig. \ref{fig:openenv}) highlights the value of information gain for robot exploration. Since there exist large free areas, the traveling to big frontiers allows the robot to spot the environment around it on the way to the assigned frontier.  

\subsection{Experimental results}
We highlight the value of our additions to the existing approaches by comparing against the nearest frontiers \cite{yamauchi_decentralized_1999}, nextFrontier \cite{colares2016next} and minPos \cite{bautin_minpos_2012} algorithms. To increase comparability between approaches we extended them to use the same general metrics for distance computation and information gain, with wavefront propagation and frontier sizing respectively. We adjusted nextFrontier, since parts of the algorithm negatively degraded the performance with our robot setup, we removed the added frontier weighting based on unknown squares around it. Our evaluation specifically focuses on exploration time taken to explore a given environment with groups of robots ranging from 2 to 5. All performance metrics are solely reliant on time taken to explore an environment, since time taken and travel distance were strongly linked for our settings. Because of the adjustments we made to other algorithms, all algorithms have a very similar way of being computed without violating their original proposed versions. We evaluated two versions of our algorithm with a different weighting on the metrics as shown in Alg. \ref{alg: main}; coExplore relies more on the visibility potential of frontiers, while co122 places an emphasize on exploring nearest frontiers. 

The results for the maze environment are visualized in Fig. \ref{fig:maze_results}. CoExplore performs best until 4 robots are used, while co122 has comparable performance to the other algorithms but improves greatly compared to coExplore for the 4 and 5 robot setting. For the experiments performed in the office environment \ref{fig:office_results}, co122 has the best performance overall. NextFrontier has comparable performance but falls off slightly for larger robot numbers, even being overtaken by coExplore for 4+ robots. Lastly for the open environment \ref{fig:open_results}, coExplore outperforms all other algorithms consistently, while co122 and nextFrontier have similar performance except for the 2 robot case. The results can be explained by looking at the information the algorithms use to compute the frontier assignment. The algorithms of Yamauchi et al. \cite{yamauchi_decentralized_1999} and Bautin et al. \cite{bautin_minpos_2012} perform poorly in an open environment where the usage of information gain greatly improves the exploration capabilities as visible by the large difference in exploration times. Performance differences between coExplore and co122 on that map can be explained in the same way. Compared to the next best algorithm, nextFrontier, our algorithms (coExplore and co122) outperform by 5 and 3\% overall, with 5 and 6\% in the structured environments and 2 and -5 in the open space. While the performance for structured environments is similar, the way they are obtained differs greatly. Co122 gains the advantage by outperforming consistently while coExplore outperforms on the maze environment by 10\% but only matches the performance in the office environment. We recommend the use of coExplore if the environments are unknown, in structured environments we recommend co122 because of the added consistency. The performance is gained by finding a balance between valuing the implicit gain of exploring a frontier through its size, identifying the frontiers closest to each robot and the frontiers closer to other robots which lead to a better spread in the environment. Different algorithms will perform better in different scenarios with stronger emphasis placed on frontier size aiding in open environments whereas decreasing that emphasis leads to better performance in more structured environments, since the need for backtracking is minimized.

Even with similar algorithms and setups, the performance increased by 5\%, which further validates that adding multiple metrics during exploration can improve performance. Since the algorithms have no reliable upper bound for exploration performance given an environment, robot setup and number of robots, we can not be sure of how good the performance is compared to the optimal solution per environment.

\section{Conclusion}\label{chapter: conclusion}
We showed a new addition to the multi-robot exploration task, where a set of robots explores a previously unknown environment to completion, i.e. total coverage of the coverable area. Our method extends previous work (minPos \cite{bautin_minpos_2012}) by combining known metrics like frontier distance and frontier information gain in order to benefit from the combined information. This enables us to explore all tested environments with comparable speed and outperforming the competition for overall exploration time.

We evaluated the performance of our algorithm with up to 5 robots in closed structured and simple open environments. Overall coExplore outperformed the next closest competition by 5\% through all scenarios and co122 by 2\%. We further demonstrate, that our approach has similar performance in urban and closed settings in all metrics. This highlights the validity of the introduction of information gain in the assignment metric.

Beyond that, we provide a ready-to-use environment for multi-robot exploration to lower the barrier to entry for research in this field. The included evaluation can be used to test ideas without needing to worry about the experimentation design, setup or logging capabilities.

\section{Future work}\label{chapter: future work}
Both, our exploration algorithm as well as the proposed testing environment, have multiple areas which can be improved upon. 
One of the problems with current multi-robot exploration researches is the unknown of what an optimal solution is, given a certain environment and settings. One future possible way to find this, is specifically overfitting deep or reinforcement learning techniques for exploration, and training them exclusively on one setting. That way a soft 'ceiling' for algorithmic performance can be found for which to aim. And looking at the assignment of frontiers at specific points, new techniques can be developed which rely on that matching. 

Because of computational constraints further optimizations and evaluations can be made to our algorithm. The algorithmic performance was evaluated on 3 mid-sized maps with up to 5 robots. This can be extended with more resources to greater amounts of robots and larger environments which can derive the potential scalability of our approach.

Another option would be the development of specific simulation tools, specializing in the construction and evaluation of multi-robot exploration scenarios which allow for faster and more reliable verification of algorithmic performance in the absence of noise.






\bibliographystyle{IEEEtran}
\bibliography{bibliography}

\end{document}